\documentclass[conference]{IEEEtran}
\IEEEoverridecommandlockouts

\usepackage{cite}
\usepackage{amsmath,amssymb,amsfonts}
\usepackage{algorithmic}
\usepackage{graphicx}
\usepackage{textcomp}
\usepackage{booktabs}
\usepackage{xcolor}
\usepackage{multirow}
\usepackage{tabularx}
\def\BibTeX{{\rm B\kern-.05em{\sc i\kern-.025em b}\kern-.08em
    T\kern-.1667em\lower.7ex\hbox{E}\kern-.125emX}}
\begin{document}
\title{Col-OLHTR: A Novel Framework for Multimodal Online Handwritten Text Recognition
\thanks{$^\dag$ Corresponding author.}
}
\author{\IEEEauthorblockN{Chenyu Liu\textsuperscript{1, 2} ~~ Jinshui Hu\textsuperscript{2} ~~ Baocai Yin\textsuperscript{2} ~~Jia Pan\textsuperscript{2} ~~ Bing Yin\textsuperscript{2} ~~ Jun Du\textsuperscript{1, $\dag$} ~~ Qingfeng Liu\textsuperscript{1, 2}}
\IEEEauthorblockA{\textsuperscript{1}\textit{University of Science and Technology of China, Hefei, China} \\
\textsuperscript{2}\textit{iFLYTEK Research, Hefei, China} \\
cyliu7@mail.ustc.edu.cn, jundu@ustc.edu.cn}
}
\maketitle
\begin{abstract}
Online Handwritten Text Recognition (OLHTR) has gained considerable attention for its diverse range of applications. Current approaches usually treat OLHTR as a sequence recognition task, employing either a single trajectory or image encoder, or multi-stream encoders, combined with a CTC or attention-based recognition decoder. However, these approaches face several drawbacks: 1) single encoders typically focus on either local trajectories or visual regions, lacking the ability to dynamically capture relevant global features in challenging cases; 2) multi-stream encoders, while more comprehensive, suffer from complex structures and increased inference costs. To tackle this, we propose a Collaborative learning-based OLHTR framework, called Col-OLHTR, that learns multimodal features during training while maintaining a single-stream inference process. Col-OLHTR consists of a trajectory encoder, a Point-to-Spatial Alignment (P2SA) module, and an attention-based decoder. The P2SA module is designed to learn image-level spatial features through trajectory-encoded features and 2D rotary position embeddings. During training, an additional image-stream encoder-decoder is collaboratively trained to provide supervision for P2SA features. At inference, the extra streams are discarded, and only the P2SA module is used and merged before the decoder, simplifying the process while preserving high performance.  Extensive experimental results on several OLHTR benchmarks demonstrate the state-of-the-art (SOTA) performance, proving the effectiveness and robustness of our design.

\end{abstract}
\section{Introduction}
Online Handwritten Text Recognition (OLHTR) aims to recognize characters from sequential sensor signal inputs, playing a crucial role in information extraction and human-computer interaction. With advancements in deep learning \cite{alexnet, resnet, transformer} and sequence modeling \cite{CTC, NMT, LAS, Coverage}, recent OLHTR systems \cite{frinken2015blstm, du2016deep, keysers2016multi, xie2017learning, du2017writer, carbune2020fast, ismail2020inceptiontime, GLRNET, xu2024multi} have seen significant improvements, benefiting numerous real-world applications.

Existing deep learning-based methods typically treat OLHTR as a sequence-to-sequence problem, using a CNN or RNN-based encoder for sequence feature extraction and an RNN or Transformer-based decoder for sequence recognition and language modeling. According to the way of sequence feature extraction, current OLHTR methods can be further divided into three categories: trajectory-based, image-based, and multimodal fusion-based.

Trajectory-based approaches usually convert sensor signals into sequential trajectory points, which serve as inputs to various neural networks for recognition. Researchers have designed different trajectory features and network architectures to enhance feature extraction and improve performance. \cite{graves2008novel} introduced a set of point-based features, including coordinates, pen state, angle, and distance, which proved effective and have been widely adopted in subsequent studies. Additionally, \cite{frinken2015blstm} utilized BLSTM networks for feature extraction, while \cite{TCRN} introduced a novel temporal convolutional recurrent network (TCRN), achieving state-of-the-art performance. Since then, TCRN has become the default network architecture for trajectory-based approaches. However, trajectory-based approaches struggle to directly capture spatial features, leading to limitations when handling complex tasks such as sloppy writing styles or Chinese character recognition.

Image-based methods usually convert sensor signals into 2D images by drawing lines on an empty background, then treating OLHTR as an offline image recognition problem \cite{wang2020writer, tanaka2021text, peng2022recognition}. However, these approaches overlook the temporal dynamics present in the original sensor signals, which trajectory-based methods naturally capture. As a result, recent studies have explored hybrid approaches that combine both image and trajectory features to improve performance in OLHTR tasks.

Multimodal fusion-based OLHTR approaches typically follow two paradigms: early fusion methods, which involve designing bimodal input features, and late fusion methods, which operate under a multi-stream framework. Early fusion methods focus on creating bimodal handcrafted features. For instance, \cite{graves2008novel} introduced region-level handcrafted image features and combined them with trajectory features as model inputs. \cite{eight} developed 8-directional images to encode online information, while \cite{gao2021handwritten, chen2017compact, xie2017learning} proposed the use of path signature \cite{pathsig} images, which better model both online and offline raw features. Late fusion methods, with their multi-stream architectures, generally achieve better performance. \cite{twostreamIndic, twostreamHMER} proposed separate online and offline encoders with a multimodal fusion block, and the recent Multi-scale Bi-Fusion framework \cite{xu2024multi} employs a multi-scale, multi-stream fusion approach using Transformers \cite{transformer}, resulting in significant performance improvements.

\begin{figure*}[ht!]
\centerline{\includegraphics[width=0.92\linewidth]{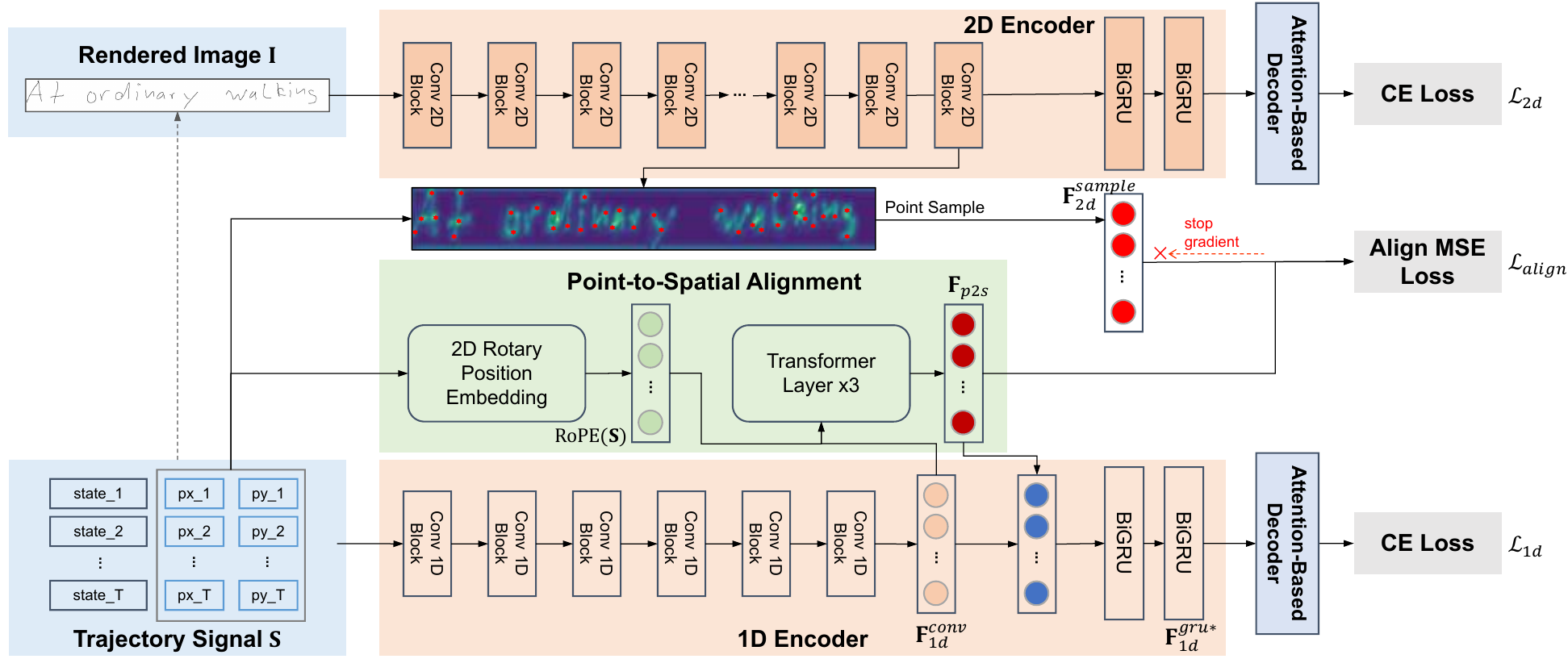}}
\caption{Overall training framework of the proposed Col-OLHTR. During testing, only the bottom trajectory stream and the P2SA module are utilized.}
\label{fig}
\end{figure*}
Despite the state-of-the-art performance of multi-stream OLHTR approaches, their complexity remains a key drawback due to extra streams, additional fusion blocks, and the need for image rendering. To address this, we propose a collaborative learning framework, Col-OLHTR. Our motivation is to combine the strengths of both trajectory-based and multi-stream methods: (1) trajectory-based methods have a simple structure but lack the ability to capture spatial features in more challenging cases; (2) multi-stream methods, while more complex, can effectively extract and utilize both sequential points and spatial image features. Inspired by this, we aim to build an OLHTR system with two key characteristics: the ability to learn multi-stream features and a simplified inference process that uses only one stream. To achieve this, Col-OLHTR introduces the P2SA module, which extracts image-level features from point positions and trajectory features, and employs a multi-stream collaborative learning framework with single-stream inference. Our contributions are as follows:

\begin{enumerate}
\item We propose a novel collaborative learning framework for OLHTR that explicitly learns multimodal features during training while maintaining a simple single-stream inference process;
\item The Point-to-Spatial Alignment (P2SA) module with its training strategy are meticulously designed to enable an effective feature learning and mapping mechanism;
\item Col-OLHTR achieves state-of-the-art (SOTA) performance on various OLHTR benchmarks, demonstrating the effectiveness and robustness of our approach in learning bimodal features while preserving simplicity.
\end{enumerate}

\section{Methodology}
\subsection{Overall Architecture}
Given a sequence of handwriting trajectory signals $\mathbf{S}=\{S_{t}=(px_t, py_t, s_t)|t=1,2,\cdots,T\}$, where $px_t, py_t$ and $s_t$ denote the writing positions and pressing state at each time step, the goal of Online Handwritten Text Recognition (OLHTR) is to generate a corresponding sequence of characters $\mathbf{y}=\{y_t |t=1,2,\cdots,L\}$.

The framework of the proposed Col-OLHTR is illustrated in Fig. \ref{fig}. Col-OLHTR utilizes multi-stream collaborative learning during training and single-stream inference during testing. 
Specifically, during the training stage, there are two main streams: the trajectory-level and the image-level streams. In the trajectory-level stream, $\mathbf{S}$ is directly fed into a 1D encoder and an attention-based decoder, where a stack of 1D convolution and GRU \cite{GRU} layers is used. In the image-level stream, $\mathbf{S}$ is first rendered into a 2D image $\mathbf{I}$ by drawing lines on an empty background. Then, $\mathbf{I}$ is fed into a 2D encoder and an attention-based decoder, where a stack of 2D convolution and GRU layers is used. Furthermore, to enable spatial perception from trajectory-level features, the Point-to-Spatial Alignment (P2SA) module and the collaborative learning paradigm are introduced, which are explained in detail in Section \ref{sec:P2SA}.
During the inference stage, only the trajectory-level stream and the P2SA module are used, significantly reducing inference costs compared to existing multimodal fusion methods.
\subsection{Encoders}
\subsubsection{1D Encoder} 
The architecture of our 1D encoder is similar to TCRN \cite{TCRN}, where 6 layers of 1D convolution with a total stride of 8 and 2 layers of BiGRU are employed to encode the raw trajectory $\mathbf{S}$ into a temporally downsampled feature $\mathbf{F}_{1d}^{gru} \in \mathbb{R}^{\frac{T}{8}\times d}$.

Mathematically, given $\mathbf{S}\in \mathbb{R}^{T\times 3}$, we firstly extract the feature in local temporal level:
\begin{equation}
\mathbf{F}_{1d}^{conv} = \mathrm{Conv1Ds}(\mathbf{S}),
\label{conv1d}
\end{equation}
where $\mathrm{Conv1Ds}(\cdot)$ denotes 1D Convolution layers. Then, two BiGRU layers are conducted to enhance the overall contexts:
\begin{equation}
\mathbf{F}_{1d}^{gru} = \mathrm{BiGRUs}(\mathbf{F}_{1d}^{conv}).
\label{gru1d}
\end{equation}
Both $\mathbf{F}_{1d}^{conv}$ and $\mathbf{F}_{1d}^{gru}$ are in $\mathbb{R}^{\frac{T}{8}\times d}$, where $d$ is the feature dimension of our encoder and set to 320 by default.

\subsubsection{2D Encoder}
The architecture of our 2D encoder is almost identical to that of the 1D encoder, except for the convolutional component. The input to the 2D encoder is the rendered image $\mathbf{I}\in\mathbb{R}^{32\times W}$, so we replace the 1D CNNs with 2D CNNs, utilizing a ResNet-18 with a total stride of $(32, 8)$:
\begin{equation}
\mathbf{F}_{2d}^{conv} = \mathrm{ResNet\text{-}18}(\mathbf{I}),
\label{res18}
\end{equation}
where $\mathbf{F}_{2d}^{conv}\in\mathbb{R}^{\frac{W}{8}\times d}$, and is also fed into 2 BiGRU layers:
\begin{equation}
\mathbf{F}_{2d}^{gru} = \mathrm{BiGRUs}(\mathbf{F}_{2d}^{conv}).
\label{gru2d}
\end{equation}

\subsection{Point-to-Spatial Alignment}
\label{sec:P2SA}
Our goal is to enhance spatial features and establish contextual connections beyond local trajectories without relying on image-level inputs. To achieve this, the Point-to-Spatial Alignment (P2SA) module is designed to explicitly map the trajectory-level features to spatial-level features.

The P2SA takes point-level features $\mathbf{F}_{1d}^{conv}$ together with its 2D position encoding as inputs, and uses three Transformer layers with self attentions to establish the global correspondence and get the mapped features:
\begin{equation}
\begin{aligned}
\mathbf{F}_{1d}^{pos} = & ~\mathbf{F}_{1d}^{conv} + \mathrm{RoPE}(\mathbf{S}),\\
\mathbf{F}_{p2s} = & ~\mathrm{Transformers}(\mathbf{F}_{1d}^{pos}),
\end{aligned}
\label{p2sa}
\end{equation}
where $\mathrm{RoPE}(\cdot)$ denotes a standard 2D rotary position embedding \cite{RoPE}, and $\mathrm{Transformers}(\cdot)$ represents standard Transformer encoder layers \cite{transformer}.

To make $\mathbf{F}_{p2s}$ learn the desired representation, an auxiliary loss is designed for collaborative learning of both streams and the P2SA module. Specifically, during training, each points in $\mathbf{S}$ can be strictly located in $\mathbf{I}$, i.e., all points in $\mathbf{F}_{1d}^{conv}$ or $\mathbf{F}_{p2s}$ can be located in $\mathbf{F}_{2d}^{conv}$ after a linear interpolation:
\begin{equation}
\mathbf{F}_{2d}^{sample} = \mathrm{Interpolation}(\mathbf{F}_{2d}^{conv}, \mathbf{S}).
\label{alignsample}
\end{equation}
This operation is like RoiAlign in \cite{maskrcnn}, but performs at point-level instead of region-level. Then, an MSE loss is adopted:
\begin{equation}
\mathcal{L}_{align} = \|\mathbf{F}_{p2s} - \mathrm{SG}(\mathbf{F}_{2d}^{sample})\|_2^2,
\label{loss_p2sa}
\end{equation}
where $\mathrm{SG}(\cdot)$ denotes a stop gradient operation. By merging the P2SA feature into $\mathbf{F}_{1d}^{conv}$, the feature in 1D encoder of Eqn. \eqref{gru1d} is updated as:

\begin{equation}
\mathbf{F}_{1d}^{gru*} = \mathrm{BiGRUs}(\mathbf{F}_{1d}^{conv} + \mathbf{F}_{p2s}).
\label{gru1d_new}
\end{equation}

\subsection{Decoder}
We adopt an attention-based autoregressive decoder \cite{LAS} for both streams, and it contains a single GRU layer and a cross attention layer. For a single decoding step $t$ and encoded features $\mathbf{F}_{enc}$, the decoding process is:

\begin{equation}
\begin{aligned}
y_{t-1} & = \mathrm{WordEmbed}(y_{t-1}^{*}), \\
a_{t} & = \mathrm{Attention}(y_{t-1}+s_{t-1}, \mathbf{F}_{enc}, \mathbf{F}_{enc}), \\
s_{t} & = \mathrm{GRUCell}(s_{t-1}, y_{t-1}+a_{t}), \\
p_{t} &= \mathrm{Softmax}(\mathbf{W}_o s_{t}),\\
y_{t}^{*} & = \arg\max p_{t}, 
\end{aligned}
\label{decoding_step}
\end{equation}
where $s_t$ and $a_t$ represents the GRU state and attention feature respectively, $\mathrm{Attention}(\mathrm{query},\mathrm{key},\mathrm{value})$ denotes a standard attention layer, and $\mathbf{W}_o$ denotes a learnable FC layer. For the point-level stream, $\mathbf{F}_{enc}$ is the $\mathbf{F}_{1d}^{gru*}$ in Eqn. \eqref{gru1d_new}, and for the image-level stream, $\mathbf{F}_{enc}$ is the $\mathbf{F}_{2d}^{gru}$ in Eqn. \eqref{gru2d}. The loss function for each stream is a cross entropy loss for each step:
\begin{equation}
\mathcal{L}_{\{1d, 2d\}} = \mathrm{CrossEntropy}(p_{t}^{\{1d, 2d\}}, y_{t}^{GT}).
\label{loss_ce}
\end{equation}
\subsection{Optimization and Inference}
The overall training loss of the framework is:
\begin{equation}
\mathcal{L}_{all} = \mathcal{L}_{1d} + \mathcal{L}_{2d} + \lambda \cdot \mathcal{L}_{align},
\label{loss_all}
\end{equation}
where $\lambda$ is hyper-parameter and is set to 2.0 by default.

During inference, only the point-level stream together with the P2SA module are used, the forward process is the sequence of [Eqn. \eqref{conv1d}, Eqn. \eqref{p2sa}, Eqn. \eqref{gru1d_new}, Eqn. \eqref{decoding_step}].

\section{Experiments}
\subsection{Datasets}
To verify the effectiveness of our proposed method, we conduct experiments on several OLHTR benchmark datasets: IAM-OnDB \cite{IAMOnDB}, OnHW-WordsTraj \cite{OnHW}, CASIA-OLHWDB \cite{CASIAOnline}, and ICDAR2013-Online \cite{ICDAR2013Online}.

IAM-OnDB \cite{IAMOnDB} is a widely used evaluation dataset for OLHTR. It consists of 13,049 handwritten English text lines produced by 221 writers, further divided into 5,363 lines for training, 2,956 lines for validation, and 3,859 lines for testing.

OnHW-WordsTraj \cite{OnHW} contains 16,752 lines of 4257 English words, and an official split of 5-fold cross validation is used for training and evaluation.

CASIA-OLHWDB \cite{CASIAOnline} is the most widely used online handwritten Chinese dataset, comprising CASIA-OLHWDB1.0-1.2 and CASIA-OLHWDB2.0-2.2. CASIA-OLHWDB1.0-1.2 contains 3,912,017 isolated characters from 1,020 writers, while CASIA-OLHWDB2.0-2.2 contains 52,220 text lines from 1,019 writers. Following the official splits of CASIA-OLHWDB2.0-2.2, 41,710 text lines are used for training, and 10,510 text lines are used for validation.

ICDAR2013-Online dataset \cite{ICDAR2013Online} consists of 3,432 handwritten Chinese text lines produced by 60 writers. It is particularly challenging due to its casual and sloppy writing style. Following previous work, we use it as an evaluation set for the CASIA-OLHWDB.
\subsection{Experimental Settings} 
All experiments were conducted on four 32GB Nvidia Tesla V100 GPUs with a batch size of 32. The learning rate starts at 2e-4 and decays to 2e-7 following a cosine schedule, using the Adam optimizer \cite{adam}. We set the training epochs to 50 for the CASIA-OLHWDB dataset and 100 for the IAM-OnDB and OnHW-WordsTraj datasets. During training, random perturbations are applied to 20\% of stroke points as data augmentation, and the image $\mathbf{I}$ is resized to a fixed height of 32 pixels. To ensure fair comparisons, we present results without external language models.

Following most previous works, we use the Character Error Rate (CER) and Word Error Rate (WER) to evaluate the recognition performance of English text. For Chinese character recognition, we adopt the Accuracy Rate (AR) and Correct Rate (CR) metrics as proposed in [14] for evaluation.

\subsection{Main Results}
To validate the effectiveness of our method, we conducted extensive experiments on the aforementioned datasets and compared the results with state-of-the-art (SOTA) methods. 

Based on different types of model inputs, we categorize previous methods into trajectory-based, image-based, or a combination of both, as shown in the following tables. Trajectory-based methods refer to those that use only sequential point information, such as the raw signal $\mathbf{S}$ or sequential geometric features \cite{graves2008novel}. Image-based methods utilize image-level inputs or features, such as raw images \cite{xu2024multi} or image features \cite{graves2008novel, chen2017compact}. Note that some inputs, like the Path Signature maps \cite{chen2017compact, xie2017learning}, combine both sequential points/strokes and image feature maps, which we classify as trajectory + image.
\begin{table}[t]
  \caption{Results on the IAM-OnDB test set.}
  \label{tab:IAM-OnDB}
  \centering
  \scriptsize
  \begin{tabular}{l|c|cc}
    \toprule
    Method & Feature & CER[\%] & WER[\%] \\  
    \midrule
    Frinken et al. BLSTM \cite{frinken2015blstm} & Trajectory & 12.30 & 25.00 \\
    Graves et al. BLSTM \cite{graves2008novel} & Trajectory+Image & 11.50 & 20.30 \\
    Keysers et al. BiLSTM \cite{keysers2016multi} & Trajectory+Image & 8.80 & 26.70 \\
    GLRNet \cite{GLRNET} & Trajectory & 8.27 & 24.22 \\
    CNN+BiLSTM \cite{benchmarking} & Trajectory & 6.94 & - \\
         Liwicki et al. LSTM \cite{liwicki2011combining}  & Trajectory+Image & - & 18.90 \\
    Carbune et al. LSTM \cite{carbune2020fast} & Trajectory & 5.90 & 18.60 \\
    Multi-scale Bi-Fusion \cite{xu2024multi} & Trajectory+Image & 4.70 & 18.10 \\
    \midrule
	\textbf{Ours} & Trajectory & \textbf{4.34} & \textbf{17.23} \\
	\bottomrule
  \end{tabular}
\end{table}

\subsubsection{Comparison on the IAM-OnDB dataset} As depicted in Tab. \ref{tab:IAM-OnDB}, SOTA works typically use both trajectory and image-level features as model inputs. The recent work, Multi-scale Bi-Fusion \cite{xu2024multi}, introduces a multi-scale, multi-stream fusion approach, achieving the best performance with a CER of 4.7\%. As shown, with only raw trajectory inputs, Col-OLHTR achieves a CER of 4.34\% and a WER of 17.23\%, delivering performance comparable to the SOTA multi-stream method \cite{xu2024multi}. This demonstrates the effectiveness of our approach.
\begin{table}[t]
  \caption{Results on the OnHW-WordsTraj dataset.}
  \label{tab:OnHW-WordsTraj}
  \centering
  \scriptsize
  \begin{tabular}{l|c|cc}
    \toprule
    Method & Feature & CER[\%] & WER[\%] \\  
    \midrule
    Attention-based model \cite{benchmarking} & Trajectory & 5.78 & 33.50 \\
    GLRNet \cite{GLRNET} & Trajectory & 7.55 & 36.76 \\
    InceptionTime \cite{ismail2020inceptiontime} & Trajectory & 2.14 & 12.32 \\
    CNN+BiLSTM \cite{benchmarking} & Trajectory & 2.07 & 11.77 \\
    InceptionTime + BiLSTM \cite{ismail2020inceptiontime} & Trajectory & 2.00 & 11.34 \\
    Multi-scale Bi-Fusion \cite{xu2024multi} & Trajectory+Image & 1.78 & 11.73 \\
    \midrule
	\textbf{Ours} & Trajectory & \textbf{1.61} & \textbf{11.24} \\
	\bottomrule
  \end{tabular}
\end{table}
\subsubsection{Comparison on the OnHW-WordsTraj dataset} Tab. \ref{tab:OnHW-WordsTraj} presents results on OnHW-WordsTraj, which features significantly shorter sequence lengths compared to IAM-OnDB. Our method achieves the lowest CER of 1.61\% and WER of 11.24\%, reaching performance comparable to SOTA methods. These results demonstrate the robustness of the proposed modules for text recognition across different scales.

\subsubsection{Comparison on the ICDAR2013-Online dataset} To verify the effectiveness of Col-OLHTR on more complex character recognition tasks, we evaluate it using the ICDAR2013-Online challenge dataset, which consists of Chinese text lines. We train our model on the CASIA-OLHWDB2.0-2.2 training set, along with synthesized samples generated from the single characters of CASIA-OLHWDB1.0-1.2. As shown in Tab. \ref{tab:ICDAR2013-Online}, our method achieves the highest AR and CR, at 95.02\% and 95.34\%, respectively, significantly outperforming the state-of-the-art Multi-scale Bi-Fusion \cite{xu2024multi}, further demonstrating the effectiveness of our method on complex OLHTR tasks.

\begin{table}[t]
  \caption{Results on the ICDAR2013-Online test set.}
  \label{tab:ICDAR2013-Online}
  \centering
  \scriptsize
  \begin{tabular}{l|c|cc}
    \toprule
    Method & Feature & AR[\%] & CR[\%] \\  
    \midrule
	Sun et al. \cite{sun2016deep} & Trajectory & 89.12 & 90.18 \\
	VGG-DBLSTM \cite{chen2017compact} & Trajectory+Image & 87.49 & 87.98 \\
	CharNet-DBLSTM \cite{chen2017compact} & Trajectory+Image & 87.10 & 87.71 \\
	  GLRNet \cite{GLRNET} & Trajectory & 91.24 & 91.81 \\	
MC-FCRN \cite{xie2017learning} & Trajectory+Image & 92.86 & 93.53 \\		
    Multi-scale Bi-Fusion \cite{xu2024multi} & Trajectory+Image & 93.92 & 94.15 \\
    \midrule
	\textbf{Ours} & Trajectory & \textbf{95.02} & \textbf{95.34} \\
	\bottomrule
  \end{tabular}
\end{table}
\subsection{Ablation Study}
We further analyzed the effectiveness of each component in the proposed P2SA module, including the added Transformer layers, 2D RoPE positional encodings, collaborative alignment loss $\mathcal{L}_{align}$, and the stop-gradient training method in Eqn. \eqref{loss_p2sa}. Results are shown in Tab. \ref{tab:ablation}. First, a 1D encoder with an attention-based decoder was set as a baseline, achieving a CER of 5.69 and a WER of 19.84. In the second line, three Transformer layers from P2SA were added to this baseline, reducing the CER by only 0.2. Third, incorporating RoPE decreased the CER by 0.49, indicating that the model was beginning to capture 2D spatial features, benefiting recognition. Moreover, using only the additional $\mathcal{L}_{align}$ did not yield improvements, but when combined with the stop-gradient design on the 2D feature maps, performance improved significantly. These ablation studies highlight not only the structural benefits of P2SA but also the strength of its learning paradigm.
\begin{table}[t!]
\centering
  \scriptsize
  \caption{Ablation study of the P2SA module on IAM-OnDB dataset.} 
  \label{tab:ablation}
   \begin{tabular}{cccc|cc}
    \toprule
     Transformer & RoPE & AlignLoss & SG & CER[\%] & WER[\%] \\ 
    \midrule 
    & & &  & 5.69 & 19.84 \\
   \checkmark & & &  & 5.49 & 19.46 \\
   \checkmark & \checkmark &  & & 5.00 & 18.91 \\
   \checkmark & \checkmark & \checkmark & & 4.92 & 18.72 \\ 
      \checkmark & \checkmark & \checkmark & \checkmark & \textbf{4.34} & \textbf{17.23} \\

	\bottomrule
  \end{tabular}
\end{table}

\section{Conclusion}
In this paper, we propose a novel framework, Col-OLHTR, which explicitly learns multimodal features during training while maintaining a simplified single-stream inference process. To achieve this, we carefully designed the P2SA module along with an effective training and testing strategy. Extensive experimental results on several OLHTR datasets demonstrate SOTA performance, validating the effectiveness of our approach. In the future, we plan to extend this framework to more complex tasks, such as handwritten table and flowchart recognition.

\bibliographystyle{IEEEtran}
\bibliography{paper_reference}

\begin{thebibliography}{10}
\providecommand{\url}[1]{#1}
\csname url@samestyle\endcsname
\providecommand{\newblock}{\relax}
\providecommand{\bibinfo}[2]{#2}
\providecommand{\BIBentrySTDinterwordspacing}{\spaceskip=0pt\relax}
\providecommand{\BIBentryALTinterwordstretchfactor}{4}
\providecommand{\BIBentryALTinterwordspacing}{\spaceskip=\fontdimen2\font plus
\BIBentryALTinterwordstretchfactor\fontdimen3\font minus \fontdimen4\font\relax}
\providecommand{\BIBforeignlanguage}[2]{{%
\expandafter\ifx\csname l@#1\endcsname\relax
\typeout{** WARNING: IEEEtran.bst: No hyphenation pattern has been}%
\typeout{** loaded for the language `#1'. Using the pattern for}%
\typeout{** the default language instead.}%
\else
\language=\csname l@#1\endcsname
\fi
#2}}
\providecommand{\BIBdecl}{\relax}
\BIBdecl

\bibitem{alexnet}
A.~Krizhevsky, I.~Sutskever, and G.~E. Hinton, ``Imagenet classification with deep convolutional neural networks,'' \emph{Advances in neural information processing systems}, vol.~25, 2012.

\bibitem{resnet}
K.~He, X.~Zhang, S.~Ren, and J.~Sun, ``Deep residual learning for image recognition,'' in \emph{Proceedings of the IEEE conference on computer vision and pattern recognition}, 2016, pp. 770--778.

\bibitem{transformer}
A.~Vaswani, N.~Shazeer, N.~Parmar, J.~Uszkoreit, L.~Jones, A.~N. Gomez, {\L}.~Kaiser, and I.~Polosukhin, ``Attention is all you need,'' \emph{Advances in neural information processing systems}, vol.~30, 2017.

\bibitem{CTC}
A.~Graves, S.~Fetrn{\'a}ndez, F.~Gomez, and J.~Schmidhuber, ``Connectionist temporal classification: labelling unsegmented sequence data with recurrent neural networks,'' in \emph{Proceedings of the 23rd international conference on Machine learning}, 2006, pp. 369--376.

\bibitem{NMT}
D.~Bahdanau, K.~Cho, and Y.~Bengio, ``Neural machine translation by jointly learning to align and translate,'' \emph{arXiv preprint arXiv:1409.0473}, 2014.

\bibitem{LAS}
W.~Chan, N.~Jaitly, Q.~Le, and O.~Vinyals, ``Listen, attend and spell: A neural network for large vocabulary conversational speech recognition,'' in \emph{2016 IEEE international conference on acoustics, speech and signal processing (ICASSP)}.\hskip 1em plus 0.5em minus 0.4em\relax IEEE, 2016, pp. 4960--4964.

\bibitem{Coverage}
Z.~Tu, Z.~Lu, Y.~Liu, X.~Liu, and H.~Li, ``Modeling coverage for neural machine translation,'' \emph{arXiv preprint arXiv:1601.04811}, 2016.

\bibitem{frinken2015blstm}
V.~Frinken and S.~Uchida, ``Deep blstm neural networks for unconstrained continuous handwritten text recognition,'' in \emph{2015 13th international conference on document analysis and recognition (ICDAR)}.\hskip 1em plus 0.5em minus 0.4em\relax IEEE, 2015, pp. 911--915.

\bibitem{du2016deep}
J.~Du, Z.-R. Wang, J.-F. Zhai, and J.-S. Hu, ``Deep neural network based hidden markov model for offline handwritten chinese text recognition,'' in \emph{2016 23rd International Conference on Pattern Recognition (ICPR)}.\hskip 1em plus 0.5em minus 0.4em\relax IEEE, 2016, pp. 3428--3433.

\bibitem{keysers2016multi}
D.~Keysers, T.~Deselaers, H.~A. Rowley, L.-L. Wang, and V.~Carbune, ``Multi-language online handwriting recognition,'' \emph{IEEE transactions on pattern analysis and machine intelligence}, vol.~39, no.~6, pp. 1180--1194, 2016.

\bibitem{xie2017learning}
Z.~Xie, Z.~Sun, L.~Jin, H.~Ni, and T.~Lyons, ``Learning spatial-semantic context with fully convolutional recurrent network for online handwritten chinese text recognition,'' \emph{IEEE transactions on pattern analysis and machine intelligence}, vol.~40, no.~8, pp. 1903--1917, 2017.

\bibitem{du2017writer}
J.~Du, J.-F. Zhai, and J.-S. Hu, ``Writer adaptation via deeply learned features for online chinese handwriting recognition,'' \emph{International Journal on Document Analysis and Recognition (IJDAR)}, vol.~20, pp. 69--78, 2017.

\bibitem{carbune2020fast}
V.~Carbune, P.~Gonnet, T.~Deselaers, H.~A. Rowley, A.~Daryin, M.~Calvo, L.-L. Wang, D.~Keysers, S.~Feuz, and P.~Gervais, ``Fast multi-language lstm-based online handwriting recognition,'' \emph{International Journal on Document Analysis and Recognition (IJDAR)}, vol.~23, no.~2, pp. 89--102, 2020.

\bibitem{ismail2020inceptiontime}
H.~Ismail~Fawaz, B.~Lucas, G.~Forestier, C.~Pelletier, D.~F. Schmidt, J.~Weber, G.~I. Webb, L.~Idoumghar, P.-A. Muller, and F.~Petitjean, ``Inceptiontime: Finding alexnet for time series classification,'' \emph{Data Mining and Knowledge Discovery}, vol.~34, no.~6, pp. 1936--1962, 2020.

\bibitem{GLRNET}
D.~Peng, C.~Xie, H.~Li, L.~Jin, Z.~Xie, K.~Ding, Y.~Huang, and Y.~Wu, ``Towards fast, accurate and compact online handwritten chinese text recognition,'' in \emph{Document Analysis and Recognition--ICDAR 2021: 16th International Conference, Lausanne, Switzerland, September 5--10, 2021, Proceedings, Part III 16}.\hskip 1em plus 0.5em minus 0.4em\relax Springer, 2021, pp. 157--171.

\bibitem{xu2024multi}
Z.~Xu, Z.~Chen, Y.~Wu, H.~Li, W.~Lv, L.~Jin, and Q.~Wang, ``A multi-scale bimodal fusion network for robust and accurate online handwriting recognition,'' in \emph{ICASSP 2024-2024 IEEE International Conference on Acoustics, Speech and Signal Processing (ICASSP)}.\hskip 1em plus 0.5em minus 0.4em\relax IEEE, 2024, pp. 6460--6464.

\bibitem{graves2008novel}
A.~Graves, M.~Liwicki, S.~Fern{\'a}ndez, R.~Bertolami, H.~Bunke, and J.~Schmidhuber, ``A novel connectionist system for unconstrained handwriting recognition,'' \emph{IEEE transactions on pattern analysis and machine intelligence}, vol.~31, no.~5, pp. 855--868, 2008.

\bibitem{TCRN}
J.~Gan, W.~Wang, and K.~Lu, ``In-air handwritten chinese text recognition with temporal convolutional recurrent network,'' \emph{Pattern Recognition}, vol.~97, p. 107025, 2020.

\bibitem{wang2020writer}
Z.-R. Wang, J.~Du, and J.-M. Wang, ``Writer-aware cnn for parsimonious hmm-based offline handwritten chinese text recognition,'' \emph{Pattern Recognition}, vol. 100, p. 107102, 2020.

\bibitem{tanaka2021text}
R.~Tanaka, K.~Osada, and A.~Furuhata, ``Text-conditioned character segmentation for ctc-based text recognition,'' in \emph{Document Analysis and Recognition--ICDAR 2021: 16th International Conference, Lausanne, Switzerland, September 5--10, 2021, Proceedings, Part III 16}.\hskip 1em plus 0.5em minus 0.4em\relax Springer, 2021, pp. 142--156.

\bibitem{peng2022recognition}
D.~Peng, L.~Jin, W.~Ma, C.~Xie, H.~Zhang, S.~Zhu, and J.~Li, ``Recognition of handwritten chinese text by segmentation: a segment-annotation-free approach,'' \emph{IEEE Transactions on Multimedia}, vol.~25, pp. 2368--2381, 2022.

\bibitem{eight}
Z.-L. Bai and Q.~Huo, ``A study on the use of 8-directional features for online handwritten chinese character recognition,'' in \emph{Eighth International Conference on Document Analysis and Recognition (ICDAR'05)}.\hskip 1em plus 0.5em minus 0.4em\relax IEEE, 2005, pp. 262--266.

\bibitem{gao2021handwritten}
L.~Gao, H.~Zhang, and C.-L. Liu, ``Handwritten text recognition with convolutional prototype network and most aligned frame based ctc training,'' in \emph{Document Analysis and Recognition--ICDAR 2021: 16th International Conference, Lausanne, Switzerland, September 5--10, 2021, Proceedings, Part I 16}.\hskip 1em plus 0.5em minus 0.4em\relax Springer, 2021, pp. 205--220.

\bibitem{chen2017compact}
K.~Chen, L.~Tian, H.~Ding, M.~Cai, L.~Sun, S.~Liang, and Q.~Huo, ``A compact cnn-dblstm based character model for online handwritten chinese text recognition,'' in \emph{2017 14th IAPR international conference on document analysis and Recognition (ICDAR)}, vol.~1.\hskip 1em plus 0.5em minus 0.4em\relax IEEE, 2017, pp. 1068--1073.

\bibitem{pathsig}
K.-T. Chen, ``Integration of paths--a faithful representation of paths by noncommutative formal power series,'' \emph{Transactions of the American Mathematical Society}, vol.~89, no.~2, pp. 395--407, 1958.

\bibitem{twostreamIndic}
A.~K. Bhunia, S.~Mukherjee, A.~Sain, A.~K. Bhunia, P.~P. Roy, and U.~Pal, ``Indic handwritten script identification using offline-online multi-modal deep network,'' \emph{Information Fusion}, vol.~57, pp. 1--14, 2020.

\bibitem{twostreamHMER}
J.~Wang, J.~Du, J.~Zhang, and Z.-R. Wang, ``Multi-modal attention network for handwritten mathematical expression recognition,'' in \emph{2019 International Conference on Document Analysis and Recognition (ICDAR)}.\hskip 1em plus 0.5em minus 0.4em\relax IEEE, 2019, pp. 1181--1186.

\bibitem{GRU}
J.~Chung, C.~Gulcehre, K.~Cho, and Y.~Bengio, ``Empirical evaluation of gated recurrent neural networks on sequence modeling,'' \emph{arXiv preprint arXiv:1412.3555}, 2014.

\bibitem{RoPE}
J.~Su, M.~Ahmed, Y.~Lu, S.~Pan, W.~Bo, and Y.~Liu, ``Roformer: Enhanced transformer with rotary position embedding,'' \emph{Neurocomputing}, vol. 568, p. 127063, 2024.

\bibitem{maskrcnn}
K.~He, G.~Gkioxari, P.~Doll{\'a}r, and R.~Girshick, ``Mask r-cnn,'' in \emph{Proceedings of the IEEE international conference on computer vision}, 2017, pp. 2961--2969.

\bibitem{IAMOnDB}
M.~Liwicki and H.~Bunke, ``Iam-ondb-an on-line english sentence database acquired from handwritten text on a whiteboard,'' in \emph{Eighth International Conference on Document Analysis and Recognition (ICDAR'05)}.\hskip 1em plus 0.5em minus 0.4em\relax IEEE, 2005, pp. 956--961.

\bibitem{OnHW}
F.~Ott, M.~Wehbi, T.~Hamann, J.~Barth, B.~Eskofier, and C.~Mutschler, ``The onhw dataset: Online handwriting recognition from imu-enhanced ballpoint pens with machine learning,'' \emph{Proceedings of the ACM on Interactive, Mobile, Wearable and Ubiquitous Technologies}, vol.~4, no.~3, pp. 1--20, 2020.

\bibitem{CASIAOnline}
C.-L. Liu, F.~Yin, D.-H. Wang, and Q.-F. Wang, ``Casia online and offline chinese handwriting databases,'' in \emph{2011 international conference on document analysis and recognition}.\hskip 1em plus 0.5em minus 0.4em\relax IEEE, 2011, pp. 37--41.

\bibitem{ICDAR2013Online}
F.~Yin, Q.-F. Wang, X.-Y. Zhang, and C.-L. Liu, ``Icdar 2013 chinese handwriting recognition competition,'' in \emph{2013 12th international conference on document analysis and recognition}.\hskip 1em plus 0.5em minus 0.4em\relax IEEE, 2013, pp. 1464--1470.

\bibitem{adam}
D.~P. Kingma and J.~Ba, ``Adam: A method for stochastic optimization,'' \emph{arXiv preprint arXiv:1412.6980}, 2014.

\bibitem{benchmarking}
F.~Ott, D.~R{\"u}gamer, L.~Heublein, T.~Hamann, J.~Barth, B.~Bischl, and C.~Mutschler, ``Benchmarking online sequence-to-sequence and character-based handwriting recognition from imu-enhanced pens,'' \emph{International Journal on Document Analysis and Recognition (IJDAR)}, vol.~25, no.~4, pp. 385--414, 2022.

\bibitem{liwicki2011combining}
M.~Liwicki, H.~Bunke, J.~A. Pittman, and S.~Knerr, ``Combining diverse systems for handwritten text line recognition,'' \emph{Machine vision and applications}, vol.~22, pp. 39--51, 2011.

\bibitem{sun2016deep}
L.~Sun, T.~Su, C.~Liu, and R.~Wang, ``Deep lstm networks for online chinese handwriting recognition,'' in \emph{2016 15th international conference on frontiers in handwriting recognition (icfhr)}.\hskip 1em plus 0.5em minus 0.4em\relax IEEE, 2016, pp. 271--276.

\end{thebibliography}

\end{document}